\documentclass[a4paper]{article}

\usepackage{INTERSPEECH2021}
\usepackage{hyperref}
\usepackage{multirow}
\graphicspath{figures/}
\usepackage{xcolor}
\usepackage{mathtools, nccmath}
\DeclarePairedDelimiter\floor{\lfloor}{\rfloor}

\title{Incorporating External POS Tagger for Punctuation Restoration}
\name{Ning Shi$^1$, Wei Wang$^1$, Boxin Wang$^2$, Jinfeng Li$^1$, Xiangyu Liu$^1$, Zhouhan Lin$^{*3}$\thanks{$^*$ Corresponding Author.}}
\address{
  $^1$Alibaba Group, China \\
  $^2$University of Illinois at Urbana-Champaign, US \\
  $^3$Shanghai Jiao Tong University, China}
\email{\{shining.shi, luyang.ww, jinfengli.ljf, eason.lxy\}@alibaba-inc.com,  boxinw2@illinois.edu, lin.zhouhan@gmail.com}

\begin{document}

\maketitle
\begin{abstract}
Punctuation restoration is an important post-processing step in automatic speech recognition. Among other kinds of external information, part-of-speech (POS) taggers provide informative tags, suggesting each input token's syntactic role, which has been shown to be beneficial for the punctuation restoration task. In this work, we incorporate an external POS tagger and fuse its predicted labels into the existing language model to provide syntactic information. Besides, we propose sequence boundary sampling (SBS) to learn punctuation positions more efficiently as a sequence tagging task. Experimental results show that our methods can consistently obtain performance gains and achieve a new state-of-the-art on the common IWSLT benchmark. Further ablation studies illustrate that both large pre-trained language models and the external POS tagger take essential parts to improve the model's performance.
\end{abstract}

\section{Introduction}

\begin{figure}[t]
  \setlength{\abovecaptionskip}{1pt}
  \centering
  \includegraphics[width=0.77\linewidth]{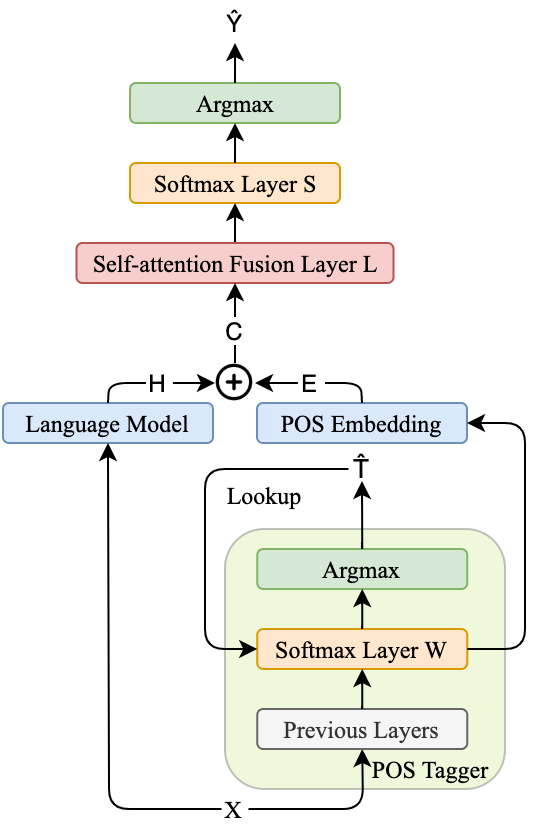}
  \caption{An illustration of our method. A token sequence $X$ of length $n$ is fed to both a LM (left) and a POS tagger (right). The POS tagger produces a sequence of predicted tags $\hat{T}$ of length $n$. To incorporate $\hat{T}$ into the LM for subsequent steps, we utilize the softmax layer weights $W \in \mathbb{R}^{b \times e}$. Elements in $\hat{T}$ serve as indexes to lookup for the corresponding columns in $W$, and form a POS embedding $E \in \mathbb{R}^{n \times b}$. The concatenation $C \in \mathbb{R}^{ n \times (d+b)}$ of both the LM hidden states $H \in \mathbb{R}^{n \times d}$ and $E$ is then forwarded to a self-attention layer to fuse the two sources of information. The eventual output is a punctuation tag sequence $\hat{Y}$ of length n to assign a punctuation tag for each token in $X$.}
  \label{fig:fusion}
  \vspace{-0.5cm}
\end{figure}

Punctuation restoration is one of the many post-processing steps in automatic speech recognition (ASR) that are non-trivial to be dealt with. At the meantime, it plays a vital role in improving the readability of the original ASR predicted speech transcripts. Huge efforts have been devoted to investigate better model structures to recover punctuation from raw lexical ASR output, including multi-layer perceptron (MLP) \cite{che2016punctuation}, conditional random field (CRF) \cite{cho2015combination}, recurrent neural networks (RNNs) \cite{tilk2015lstm,tilk2016bidirectional,Pahuja2017,wang2018self,kim2019deep}, convolutional neural networks (CNNs) \cite{che2016punctuation,wang2018self}, and transformers \cite{NIPS2017_7181,wang2018self,yi2019self,chen2020controllable}. In addition, a wide range of correlated tasks can be utilized to improve the performance of punctuation restoration via multi-task learning, such as sentence boundary detection \cite{wang2012dynamic}, capitalization recovering \cite{Pahuja2017}, disfluency removing \cite{cho2015combination}, and dependency parsing \cite{zhang2013punctuation}.

Pre-trained language models (LMs) play an increasingly important role in this task. It has been proposed to use Bidirectional Encoder Representations from Transformers (BERT) \cite{devlin-etal-2019-bert} as a building block and treat punctuation restoration as a sequence tagging task \cite{makhija2019transfer}, which significantly improved its performance. A series of follow-up works have revealed the effectiveness of other kinds of pre-trained LMs for this task \cite{courtland-etal-2020-efficient, alam-etal-2020-punctuation}.


Although it is possible to recover the punctuation merely from lexical data \cite{kim2019deep, yi2020focal}, there is other information external to it that we can utilize. Apart from the aforementioned multi-task learning schemes, multi-modality is another practical approach to fuse relevant information from different modalities, which in return leads to improved performance. By multimodal learning, prosodic cues have been proven informative in improving the quality of punctuation restoration \cite{tilk2015lstm, levy2012effect, szaszak2019leveraging,sunkara2020multimodal}. However, although bunch of works have been proposed in multimodal learning with both lexical and acoustic features, fusing part-of-speech (POS) tag knowledge into raw lexical ASR output has not been well studied yet. As one of the important information that reveals the lexical roles of each word in a sequence, POS tags are believed to be beneficial for this task. For instance, typically a sentence will not end with a definite article like ``a", ``an'', or ``the". Previous research explored using POS tag prediction as an auxiliary task in a multi-task learning scenario, and have found that such multi-task learning works for both of tasks \cite{yi2020adversarial}. Instead, we propose to explicitly use an external POS tagger to enhance textual input for punctuation restoration, enabling the model to incorporate POS tags information while learning through a more straightforward single task learning scenario.

In this work\footnote{We've made the source code and detailed evaluation results of this work publicly available at \url{ https://github.com/ShiningLab/POS-Tagger-for-Punctuation-Restoration.git}.}, we present a framework to involve POS tags provided by an external POS tagger as an extra input and combine it with a new pre-trained LM, Funnel Transformer \cite{NEURIPS2020_2cd2915e}, which effectively filters out the sequential redundancy. Our contributions are as follows:

\begin{itemize}
    \item We propose a novel framework to employ an external POS tagger to provide syntactic information for punctuation restoration, as well as a new stochastic sampling scheme called sequence boundary sampling (SBS) to better adapt to pre-trained LMs. With RoBERTa \cite{liu2019roberta}, our method sets a new state-of-the-art on IWSLT datasets in terms of Micro $F_1$.

    \item We introduce Funnel Transformer \cite{NEURIPS2020_2cd2915e} to our framework and further push the gap between our method and previous studies.

    \item As ablation study, we examine the punctuation restoration performance of a wide range of pre-trained LMs in a fair and comparable setting, which provides a wide set of pre-trained LM benchmarks on this task. 
\end{itemize}

\section{Method}

Whether a word needs to be followed by punctuation is closely related to its grammatical role. For instance, a comma is often placed before the coordinating conjunction to join two independent clauses. In this section, we introduce how we incorporate an external POS tagger for punctuation restoration. Specifically, our framework consists of a POS fusion module and the SBS batch sampling strategy.

\subsection{Fusing POS tags into LM representations}


Our model forms punctuation restoration as a sequence tagging task, which incorporates both a pre-trained LM and a trained POS tagger for final punctuation tag prediction. Assume that we have a sequence of input $X$ of length $n$, a pre-trained LM with hidden size $d$, and a neural POS tagger with hidden size $b$. Figure \ref{fig:fusion} is a visual illustration of our model.

On the LM side, we make use of the LM hidden states. Thus we can view the LM as a function $F$ parameterized by $\theta$, mapping the sequence $X$ into a sequence of context dependent embeddings $H$ by
\begin{equation}
\setlength{\abovedisplayskip}{4pt}
\setlength{\belowdisplayskip}{4pt}
H = F_{\theta}(X) \in \mathbb{R}^{n \times d},
\end{equation}
where $H$ is the last layer hidden states of the given LM. In subsequent steps, we will combine $H$ with the information from the POS tagger side.

On the POS tagger side, we leverage a neural POS tagger\footnote{In this work, we employ the fast universal POS tagging model for English \url{https://huggingface.co/flair/upos-english-fast}, but any POS tagger with its last classification layer being a softmax layer could apply.}, by using its predictions as well as its softmax layer weights. Formally, the POS tag predictions $\hat{T}$ are produced by
\begin{equation}
\hat{T} = F_W(X) \in \mathbb{R}^{n},
\end{equation}
where $F_W(\cdot)$ stands for the POS tagger, with $W \in \mathbb{R}^{b \times e}$ being its softmax layer weights. This weight matrix can be viewed as a POS embedding matrix, with each embedding having a size $b$ and $e$ being the number of POS tag classes. For simplicity, we only show the related parameters $W$ in the POS tagger. To get POS embeddings $E \in \mathbb{R}^{n \times b}$ for the input sequence, we use elements in $\hat{T}$ to lookup for the corresponding columns in $W$, and form the POS embedding $E \in \mathbb{R}^{n \times b}$. 

Further, in the fusion step, we first concatenate $H$ and $E$ alongside the sequential dimension to get a fused representation $C \in \mathbb{R}^{ n \times (d+b)}$. Different from conventional practices that combine high-level representations by concatenation alone \cite{tilk2015lstm,yi2019self}, which may suffer from the inefficacy to model the cross-modality relationship, we utilize a self-attention encoder \cite{NIPS2017_7181} as the fusion layer that enables both features to better interact with each other through the multi-head self-attention mechanism. Subsequently, we pass $C$ through the block of self-attention layer $L_{\gamma}$, as well as a final softmax layer $S_{\eta}$ to output punctuation tags $\hat{Y}$ by
\begin{equation}
\setlength{\abovedisplayskip}{4pt}
\setlength{\belowdisplayskip}{4pt}
\hat{Y} = S_{\eta}(L_{\gamma}(C)),
\end{equation}
where $\gamma$ and $\eta$ stand for the corresponding parameters in the components. Training is simply conducted as back-propagating the cross entropy loss between $\hat{Y}$ and its corresponding ground truth $Y$, over parameter $\theta, W, \gamma,$ and $\eta$, where $\theta$ is initialized from the pre-trained LM, and $W$ is initialized from the external POS tagger weights. 

\subsection{Sequence boundary sampling}

Since sentence boundaries are not explicit in raw ASR output, the raw output of the whole training set can be viewed as a continuous word stream. Due to memory constraints, it have to be truncated to align with a maximum sequence length $L$. Some previous works split the corpus into multiple sequences in preprocessing steps \cite{makhija2019transfer,alam-etal-2020-punctuation}, resulting in reusing the same truncation of the training set for every epoch. Some others rotates the training set one token at a time between different epochs \cite{courtland-etal-2020-efficient}, however yielding the training sample size too large. 
To further randomize the truncation of the continuous word stream, we propose SBS, where we uniformly select a range in the corpus $S$, starting from $x_{0 \leq j < |S| - L}$ to $x_{j+L-1}$, forming a token sequence $X=[x_j, ..., x_{j+L-1}]$ of length $L$. We limit the number of sampling times to $\floor{\frac{|S|}{L}}$ so as to maintain an acceptable training size while keeping the possibility of exposing every token at every sequence position to the model for more robust learning. This sampling mechanism provides a computationally more efficient process than earlier ways by both weakening the connection between positions and tokens and allowing mini-batches of samples to represent the entire corpus.

\section{Experiments}

\begin{table*}[t!]
\setlength{\abovecaptionskip}{2pt}
\caption{An example of pre-processed data to align with BERT (bert-base-uncased).} 
\centering
\resizebox{\textwidth}{!}{
\begin{tabular}{l}
\hline
\textbf{Raw Word Sequence} \\
\textbf{Raw Label Sequence} \\ 
\textbf{Token Sequence ($X$)} \\
\textbf{Label Sequence ($Y$)} \\ 
\textbf{POS Tag Sequence ($\hat{T}$)} \\
\textbf{Position Mask} \\
\hline
\end{tabular}

\begin{tabular}{cccccccccccccccccccccc}
\hline
 & adrian & kohler & well & we & 're & here & today & to & talk & about & the & puppet & horse &  \\
 & O & COMMA & COMMA & O & O & O & O & O & O & O & O & O & PERIOD & \\
 {$\langle$} BOS $\rangle$ & [CLS] & adrian & ko & \#\#hler & well & we & ' & re & here & today & to & talk & about & the & puppet & horse & [SEP] & $\langle$ EOS $\rangle$ \\
 & O & O & O & COMMA & COMMA & O & O & O & O & O & O & O & O & O & O & PERIOD & O \\
 & X & PROPN & X & PROPN & INTJ & PRON & X & VERB & ADV & NOUN & PART & VERB & ADP & DET & NOUN & NOUN & X \\
 & 0 & 1 & 0 & 1 & 1 & 1 & 0 & 1 & 1 & 1 & 1 & 1 & 1 & 1 & 1 & 1 & 0 \\
\hline
\end{tabular}
}
\vspace{-0.15cm}
\label{tab:data}
\end{table*}

\begin{table*}
  \setlength{\abovecaptionskip}{2pt}
  \caption{Evaluation results on \textit{Ref.} in terms of \textit{P}(\%), \textit{R}(\%), \textit{Micro F\textsubscript{1}}(\%), and \textit{Mean F\textsubscript{1}}(\%).}
  \label{tab:results0}
  \centering
  \resizebox{\textwidth}{!}{
  \begin{tabular}{llccc|ccc|ccc|cccc}
    \toprule
    \multirow{2}{*}{\textbf{Language Model}} & \multirow{2}{*}{\textbf{Modification}} & \multicolumn{3}{c}{\textit{\textbf{COMMA}}} & \multicolumn{3}{c}{\textit{\textbf{PERIOD}}} & \multicolumn{3}{c}{\textit{\textbf{QUESTION}}} & \multicolumn{4}{c}{\textit{\textbf{Overall}}} \\
    \cline{3-15}
     && \textit{P} & \textit{R} & \textit{F\textsubscript{1}} & \textit{P} & \textit{R} & \textit{F\textsubscript{1}} & \textit{P} & \textit{R} & \textit{F\textsubscript{1}} & \textit{P} & \textit{R} & \textit{Micro F\textsubscript{1}} & \textit{Mean F\textsubscript{1}} \\
    \midrule
    
    \multirow{8}{*}{None} 
    & DNN-A \cite{che2016punctuation} & 48.6 & 42.4 & 45.3 & 59.7 & 68.3 & 63.7 & - & - & - & 54.8 & 53.6 & 54.2 & 36.3 \\
    & CNN-2A \cite{che2016punctuation} & 48.1 & 44.5 & 46.2 & 57.6 & 69.0 & 62.8 & - & - & - & 53.4 & 55.0 & 54.2 & 36.3 \\
    & T-BRNN-pre \cite{tilk2016bidirectional} & 65.5 & 47.1 & 54.8 & 73.3 & 72.5 & 72.9 & 70.7 & 63.0 & 66.7 & 70.0 & 59.7 & 64.4 & 64.8 \\
    & Teacher-Ensemble \cite{yi2017distilling} & 66.2 & 59.9 & 62.9 & 75.1 & 73.7 & 74.4 & 72.3 & 63.8 & 67.8 & 71.2 & 65.8 & - & 68.4 \\
    & SAPR \cite{wang2018self} & 57.2 & 50.8 & 55.9 & $\textbf{96.7}^*$ & $\textbf{97.3}^*$ & $\textbf{96.8}^*$ & 70.6 & 69.2 & 70.3 & 78.2 & 74.4 & 77.4 & 74.3 \\
    & DRNN-LWMA-pre \cite{kim2019deep} & 62.9 & 60.8 & 61.9 & 77.3 & 73.7 & 75.5 & 69.6 & 69.6 & 69.6 & 69.9 & 67.2 & 68.6 & 69.0 \\
    & Self-attention \cite{yi2019self} & 67.4 & 61.1 & 64.1 & 82.5 & 77.4 & 79.9 & 80.1 & 70.2 & 74.8 & 76.7 & 69.6 & - & 72.9\\
    & CT-transformer \cite{chen2020controllable} & 68.8 & 69.8 & 69.3 & 78.4 & 82.1 & 80.2 & 76.0 & 82.6 & 79.2 & 73.7 & 76.0 & 74.9 & 76.2 \\
    
    \midrule
    
    \multirow{4}{*}{bert-base-uncased}
    & Transfer \cite{makhija2019transfer} & 72.1 & 72.4 & 72.3 & 82.6 & 83.5 & 83.1 & 77.4 & 89.1 & 82.8 & 77.4 & 81.7 & - & 79.4 \\
    & Adversarial \cite{yi2020adversarial} & 74.2 & 69.7 & 71.9 & 84.6 & 79.2 & 81.8 & 76.0 & 70.4 & 73.1 & 78.3 & 73.1 & - & 75.6 \\
    & FL \cite{yi2020focal} & 74.4 & 77.1 & 75.7 & 87.9 & 88.2 & 88.1 & 74.2 & 88.5 & 80.7 & 78.8 & 84.6 & 81.6 & 81.5 \\
    & Bi-LSTM \cite{alam-etal-2020-punctuation} & 71.7 & 70.1 & 70.9 & 82.5 & 83.1 & 82.8 & 75.0 & 84.8 & 79.6 & 77.0 & 76.8 & 76.9 & 77.8 \\
    & Ours: POS Fusion + SBS & 69.9 & 72.0 & 70.9 & 81.9 & 85.5 & 83.7 & 76.5 & 84.8 & 80.4 & 75.9 & 78.8 & 77.3 & 78.3 \\
    
    \midrule
    
    \multirow{3}{*}{bert-large-uncased}
    & Transfer \cite{makhija2019transfer} & 70.8 & 74.3 & 72.5 & 84.9 & 83.3 & 84.1 & 82.7 & 93.5 & 87.8 & 79.5 & 83.7 & - & 81.4 \\
    & Bi-LSTM \cite{alam-etal-2020-punctuation} & 72.6 & 72.8 & 72.7 & 84.8 & 84.6 & 84.7 & 70.0 & 91.3 & 79.2 & 78.3 & 79.0 & 78.6 & 78.9 \\
    & Pre-trained POS Fusion + SBS & 74.7 & 71.2 & 72.9 & 83.4 & 87.2 & 85.2 & 78.4 & 87.0 & 82.5 & 79.1 & 79.3 & 79.2 & 80.2 \\
    
    \midrule
    
    \multirow{3}{*}{roberta-base}
    & Aggregate \cite{courtland-etal-2020-efficient} & 76.9 & 75.4 & 76.2 & 86.1 & 89.3 & 87.7 & $\textbf{88.9}^*$ & 87.0 & 87.9 & 84.0 & 83.9 & - & 83.9 \\
    & Bi-LSTM \cite{alam-etal-2020-punctuation} & 73.6 & 75.1 & 74.3 & 84.9 & 87.6 & 86.2 & 77.4 & 89.1 & 82.8 & 79.2 & 81.5 & 80.3 & 81.1 \\
    & Ours: POS Fusion + SBS & 75.2 & 76.5 & 75.9 & 86.0 & 87.9 & 86.9 & 73.2 & 89.1 & 80.4 & 80.3 & 82.3 & 81.3 & 81.1 \\
    
    \midrule
    
    \multirow{4}{*}{roberta-large}
    & Aggregate \cite{courtland-etal-2020-efficient} & 74.3 & 76.9 & 75.5 & 85.8 & 91.6 & 88.6 & 83.7 & 89.1 & 86.3 & 81.3 & $\textbf{85.9}^*$ & - & 83.5 \\
    & Bi-LSTM \cite{alam-etal-2020-punctuation} & 76.9 & 75.8 & 76.3 & 86.8 & 90.5 & 88.6 & 72.9 & 93.5 & 81.9 & 81.6 & 83.3 & 82.4 & 82.3 \\
    & Bi-LSTM + augmentation \cite{alam-etal-2020-punctuation} & 76.8 & 76.6 & 76.7 & 88.6 & 89.2 & 88.9 & 82.7 & 93.5 & 87.8 & 82.6 & 83.1 & 82.9 & 84.5 \\
    & Ours: POS Fusion + SBS & 77.4 & 79.4 & 78.4 & 87.7 & 89.6 & 88.6 & 80.4 & 89.1 & 84.5 & 82.4 & 84.6 & 83.5 & 83.9 \\
    
    \midrule
    
    \multirow{4}{*}{funnel-transformer-xlarge}
    & None & 75.5 & $\textbf{82.4}^*$ & $\textbf{78.8}^*$ & 88.7 & 89.0 & 88.9 & 82.4 & 91.3 & 86.6 & 81.7 & 85.8 & 83.7 & 84.7 \\
    & SBS & 77.2 & 80.1 & 78.6 & 88.4 & 89.4 & 88.9 & 86.3 & $\textbf{95.7}^*$ & $\textbf{90.7}^*$ & 82.7 & 85.0 & 83.8 & $\textbf{86.1}^*$ \\
    & -POS embedding +SBS & 76.4 & 80.9 & 78.6 & 87.9 & 90.2 & 89.0 & 82.4 & 91.3 & 86.6 & 81.9 & 85.6 & 83.7 & 84.7 \\
    & POS Fusion + SBS & $\textbf{78.9}^*$ & 78.0 & 78.4 & 86.5 & 93.4 & 89.8 & 87.5 & 91.3 & 89.4 & $\textbf{82.9}^*$ & 85.7 & $\textbf{84.3}^*$ & 85.9 \\
    
    \bottomrule
  \end{tabular}
}
\end{table*}

In this section, we conduct experiments on IWSLT dataset, as well as ablation studies to investigate the efficacy of SBS, POS fusion and several pre-trained LMs respectively. Experimental results demonstrate that our proposed methods with SBS and POS fusion can achieve state-of-the-art performance on IWSLT datasets. Finally, we conduct case studies to further evaluate some succeed and failure cases.

\subsection{Experimental setup}

\begin{table*}
  \setlength{\abovecaptionskip}{2pt}
  \caption{Evaluation results on \textit{ASR} in terms of \textit{P}(\%), \textit{R}(\%), \textit{Micro F\textsubscript{1}}(\%), and \textit{Mean F\textsubscript{1}}(\%).}
  \label{tab:results1}
  \centering
  \resizebox{\textwidth}{!}{
  \begin{tabular}{llccc|ccc|ccc|cccc}
    \toprule
    \multirow{2}{*}{\textbf{Language Model}} & \multirow{2}{*}{\textbf{Modification}} & \multicolumn{3}{c}{\textit{\textbf{COMMA}}} & \multicolumn{3}{c}{\textit{\textbf{PERIOD}}} & \multicolumn{3}{c}{\textit{\textbf{QUESTION}}} & \multicolumn{4}{c}{\textit{\textbf{Overall}}} \\
    \cline{3-15}
     && \textit{P} & \textit{R} & \textit{F\textsubscript{1}} & \textit{P} & \textit{R} & \textit{F\textsubscript{1}} & \textit{P} & \textit{R} & \textit{F\textsubscript{1}} & \textit{P} & \textit{R} & \textit{Micro F\textsubscript{1}} & \textit{Mean F\textsubscript{1}} \\
    \midrule
    
    \multirow{3}{*}{None} 
    & T-BRNN-pre \cite{tilk2016bidirectional} & 59.6 & 42.9 & 49.9 & 70.7 & 72.0 & 71.4 & 60.7 & 48.6 & 54.0 & 66.0 & 57.3 & 61.4 & 58.4 \\
    & Teacher-Ensemble \cite{yi2017distilling} & 60.6 & 58.3 & 59.4 & 71.7 & 72.9 & 72.3 & 66.2 & 55.8 & 60.6 & 66.2 & 62.3 & - & 64.1 \\
    & Self-attention \cite{yi2019self} & 64.0 & 59.6 & 61.7 & 75.5 & 75.8 & 75.6 & $\textbf{72.6}^*$ & 65.9 & $\textbf{69.1}^*$ & 70.7 & 67.1 & - & 68.8 \\
    
    \midrule
    
    \multirow{3}{*}{bert-base-uncased}
    & Adversarial \cite{yi2020adversarial} & $\textbf{70.7}^*$ & 68.1 & $\textbf{69.4}^*$ & 77.6 & 77.5 & 77.5 & 68.4 & 66.0 & 67.2 & $\textbf{72.2}^*$ & 70.5 & - & $\textbf{71.4}^*$ \\
    & FL \cite{yi2020focal} & 59.0 & $\textbf{76.6}^*$ & 66.7 & 78.7 & 79.9 & 79.3 & 60.5 & 71.5 & 65.6 & 66.1 & 76.0 & 70.7 & 70.5 \\
    & Bi-LSTM \cite{alam-etal-2020-punctuation} & 49.3 & 64.2 & 55.8 & 75.3 & 76.3 & 75.8 & 44.7 & 60.0 & 51.2 & 60.4 & 70.0 & 64.9 & 61.0 \\
    & Ours: POS Fusion + SBS & 49.3 & 65.6 & 56.3 & 73.6 & 78.8 & 76.1 & 48.9 & 62.9 & 55.0 & 60.0 & 72.0 & 65.4 & 62.5 \\

    \midrule
    
    \multirow{2}{*}{bert-large-uncased}
    & Bi-LSTM \cite{alam-etal-2020-punctuation} & 49.9 & 67.0 & 57.2 & 77.0 & 78.9 & 77.9 & 50.0 & 74.3 & 59.8 & 61.4 & 73.0 & 66.7 & 65.0 \\
    & Ours: POS Fusion + SBS & 54.7 & 64.3 & 59.1 & 75.8 & 82.5 & 79.0 & 48.8 & 60.0 & 53.9 & 64.6 & 73.2 & 68.6 & 64.0 \\

    \midrule
    
    \multirow{2}{*}{roberta-base}
    & Bi-LSTM \cite{alam-etal-2020-punctuation} & 51.9 & 69.3 & 59.3 & 77.5 & 80.3 & 78.9 & 50.0 & 65.7 & 56.8 & 62.8 & 74.7 & 68.2 & 65.0 \\
    & Ours: POS Fusion + SBS & 55.5 & 68.7 & 61.4 & 78.0 & 81.1 & 79.5 & 51.1 & 68.6 & 58.5 & 65.5 & 74.8 & 69.8 & 66.5 \\

    \midrule
    
    \multirow{3}{*}{roberta-large}
    & Bi-LSTM \cite{alam-etal-2020-punctuation} & 56.6 & 67.9 & 61.8 & 78.7 & 85.3 & 81.9 & 46.6 & 77.1 & 58.1 & 66.5 & 76.7 & 71.3 & 67.3 \\
    & Bi-LSTM + augmentation \cite{alam-etal-2020-punctuation} & 64.1 & 68.8 & 66.3 & 81.0 & 83.7 & 82.3 & 55.3 & 74.3 & 63.4 & 72.0 & 76.2 & $\textbf{74.0}^*$ & 70.7 \\
    & Ours: POS Fusion + SBS & 59.6 & 68.0 & 63.5 & 79.5 & 86.0 & 82.6 & 50.0 & 77.1 & 60.7 & 68.8 & 77.0 & 72.7 & 68.9 \\
    
    \midrule
    
    \multirow{4}{*}{funnel-transformer-xlarge}
    & None & 52.6 & 76.5 & 62.3 & $\textbf{81.2}^*$ & 81.8 & 81.5 & 53.1 & 74.3 & 61.9 & 64.1 & 79.1 & 70.8 & 68.6 \\
    & SBS & 54.4 & 72.8 & 62.3 & 81.0 & 82.9 & 82.0 & 59.6 & 80.0 & 68.3 & 65.9 & 77.9 & 71.4 & 70.8 \\
    & -POS embedding +SBS & 54.8 & 73.4 & 62.8 & 80.7 & 85.3 & $\textbf{82.9}^*$ & 54.7 & $\textbf{82.9}^*$ & 65.9 & 66.0 & $\textbf{79.5}^*$ & 72.1 & 70.5 \\
    & POS Fusion + SBS & 56.6 & 71.6 & 63.2 & 79.0 & $\textbf{87.0}^*$ & 82.8 & 60.5 & 74.3 & 66.7 & 66.9 & 79.3 & 72.6 & 70.9 \\
    
    \bottomrule
  \end{tabular}
}
\end{table*}

\textbf{Dataset.} 
We evaluate our methods on transcriptions of TED Talks from IWSLT datasets \cite{federico2012overview}, which has been regarded as a standard benchmark on punctuation restoration \footnote{We download data from \url{https://github.com/xashru/punctuation-restoration}}. Following the standard evaluation protocol, we use the same splits: 2.1M words for training, 296K words for validation, 12626 words for the manual transcription test set (\textit{Ref.}), and 12822 words for the actural ASR transcription test set (\textit{ASR}) provided in  \cite{che2016punctuation}, where the wrong predicted words of ASR cause the task more challenging. There are four labels for each word, meaning which punctuation to follow behind. Specifically, (i) \textit{COMMA} is for commas, colons, and dashes; (ii) \textit{PERIOD} is for full stops, exclamation marks, and semicolons; (iii) \textit{QUESTION} is for question marks only; (iv) \textit{O} is for no punctuation.

\noindent \textbf{Preprocessing.} 
To maintain an original-to-tokenized alignment with the subword tokenization of LMs, we assign label \textit{O} for all non-tail subword pieces in a word and keep the original punctuation label only for the tail subword piece. Moreover, every tokenized sentence is also prefixed with a \textit{beginning-of-sentence} token $\langle$ BOS $\rangle$ and suffixed with an \textit{end-of-sentence} token $\langle$ EOS $\rangle$. Thanks to SBS, it is unnecessary to hold a \textit{pad} token since token sequences are always of the maximum sequence length. We maintain a position mask to filter out  non-tail subword pieces and special tokens from evaluation. A pre-processed data example after aligning with BERT is shown in Table \ref{tab:data}.

\noindent \textbf{Evaluation Metrics.}
We measure the performance in terms of precision (\textit{P}), recall (\textit{R}), and F1-score. We notice the definition of the F1-score is not consistent in previous studies, thus, to be on the same page, we report both micro F1-score (Micro $F_1$) and mean F1-score (Mean $F_1$). The latter is defined as the average $F_1$ score of \textit{COMMA}, \textit{PERIOD} and \textit{QUESTION}. The Mean $F_1$ is manually calculated for preceding approaches that did not state the definition of reported F1-score. With the exception of those clearly noted, we consider reported F1-score as Micro $F_1$ if it does not match Mean $F_1$ we calculated. Otherwise, we level it as Mean $F_1$ in our evaluation table.

\noindent \textbf{Implementation Details.} 
We use Transformers \cite{wolf-etal-2020-transformers} built by HuggingFace to tokenize raw words into subword units and explore different LMs involving BERT \cite{devlin-etal-2019-bert}, ALBERT \cite{lan2020albert}, RoBERTa \cite{liu2019roberta}, XLM-RoBERTa \cite{conneau-etal-2020-unsupervised}, and Funnel Transformer \cite{NEURIPS2020_2cd2915e}. 
In our experiment, we use the fast universal POS tagging model for English\footnote{\url{https://huggingface.co/flair/upos-english-fast}} by Flair \cite{akbik2019flair}, where $b = 512$ and $e = 20$. For our self-attention based fusion layer, in base LMs ($d < 1024$), the dimension of the feedforward network and the number of attention heads are 3072 and 8 respectively. Accordingly, in large LMs ($d \geq 1024$), they are set to 4096 and 16, respectively. The residual dropout rate is 0.1. We train on a single NVIDIA Tesla V100 with a maximum sequence length of 256 and a batch size of 8. Batches containing less than 8 samples are dropped. To avoid overfitting, we adopt early stopping \cite{prechelt1998early} with a patience of $8$ epochs for a lower validation loss. We use Adam optimizer \cite{DBLP:journals/corr/KingmaB14} with a learning rate of 5$e^{-6}$, and an $\ell_2$ gradient clipping of $5.0$ \cite{10.5555/3042817.3043083}. We choose to minimize a cross entropy loss instead of focal loss \cite{yi2020focal} for a fair comparison with previous results that do not make use of focal loss.

\subsection{Results}
Our results on reference text and on ASR outputs are shown in Table \ref{tab:results0} and \ref{tab:results1}, respectively. Since pre-trained LMs play a vital role in the performances, reported results are grouped according to their LMs. In each group, we list the complete version of our method, i.e., with POS tag fusion and SBS, and compare that to the corresponding published baselines using the same pre-trained LM. Our models are denoted as \emph{Ours: POS fusion + SBS} in each of the groups.

The last group of each table is named \textit{funnel-transformer-xlarge}. In this group, we provide our method using Funnel Transformer \cite{NEURIPS2020_2cd2915e} as its LM. Since there are no preceding baselines to compare, it is a good testbed setting for our method. We conduct four ablation studies with Funnel Transformer to show the relative contributions of each component in our method. We denote the base setting, which consists of only an LM and a linear layer, as \textit{None}, and \textit{SBS} for this base setting with SBS during training. We refer to \textit{-POS embedding +SBS} as our model involving POS tags as part of inputs but with POS embedding initialized with random noise. The final line denoted by \textit{POS Fusion + SBS} is the full setting of our method.


For punctuation restoration on the reference test set, we report the evaluation results in Table \ref{tab:results0}. Firstly, our method in the group of RoBERTa (\textit{roberta-large}) outperforms all the previous models in terms of the overall Micro $F_1$ and on-par with them in terms of Mean $F_1$. Further incorporating Funnel Transformer, our model achieves the new state-of-the-art, resulting in an absolute improvement of $1.4\%$ in both Micro $F_1$ and Mean $F_1$ compared to previous best \cite{alam-etal-2020-punctuation}. Note that this is without focal loss and data augmentation, which suggests that it can be further pushed up. Compared with \cite{alam-etal-2020-punctuation}, our method consistently improves performances in all three individual classes.

For punctuation restoration on the \textit{ASR} test set, we report the evaluation results in Table \ref{tab:results1}. Without using data augmentation, our \textit{funnel-transformer-xlarge} based model obtains $72.6\%$ in Micro $F_1$ and $70.9\%$ in Mean $F_1$, outperforming previous best \textit{roberta-large} based model \cite{alam-etal-2020-punctuation} by absolute $1.3\%$ and $3.6\%$ separately. In terms of Mean $F_1$, our final version with Funnel Transformer achieves competitive results, the only previous one that outperforms our method is through adversarial learning \cite{yi2020adversarial}. Moreover, the aforementioned trend on \textit{Ref.} can also be noticed on \textit{ASR} as well, including the positive impact of large LMs, \textit{SBS}, and \textit{POS Fusion} with POS tagger provided POS embeddings. Compared to the \textit{ref} test set, the primary extra difficulty in \textit{ASR} test set is the noise caused by the incorrectly predicted words. Since POS is an attribute of the natural language, our method thus exhibits heavier dependency on the preceding ASR outputs. An incorrect word will likely be assigned with an incorrect POS tag, leading to a misrepresentation of both lexical and POS features. It can be seen that data augmentation techniques that simulate error words created by the ASR system play an essential role in handling the noisier test set. We believe our model can also benefit from data augmentation, but we leave it for future work. 

As for ablation studies, we analyze different settings within the Funnel Transformer group. Firstly, the base setting, which is Funnel Transformer simply followed by a liner layer, is already performing satisfyingly well, especially for \textit{COMMA}. With SBS, the $F_1$ score for \textit{QUESTION} goes up from $86.6\%$ to $90.7\%$ without too much loss on the other two classes. This indicates that SBS does help to alleviate the class imbalance issue. Particularly in training samples generated by SBS, \textit{PERIOD} and \textit{QUESTION} no longer frequently appear near the end of the sequence. Thus models have to focus on contexts rather than positions to infer punctuations.
In contrast to \textit{-POS embedding +SBS} equipped with a random initialized POS embedding, our full setting \textit{POS Fusion + SBS} performes the best across all settings. One possible reason is that pre-trained weights, instead of random weights, can serve as regularization or constraints for models to focus on major features from training samples and accelerate training before overfitting. We want to stress that it is stated in former studies \cite{yi2020focal,courtland-etal-2020-efficient} that base LMs are preferred due to the small size of IWSLT datasets. However, our findings suggest the opposite. 

\section{Conclusion}

We propose a novel framework that brings POS knowledge via a self-attention based fusion layer for punctuation restoration. Experiments conducted on IWSLT datasets prove that incorporating POS tags makes it possible for prior lexical-based approaches to earn significant performance gains. We also introduce a new sampling technique, SBS, that makes fuller use of the corpus and better adapts to LMs. Empirical results show that our method with Funnel Transformer is superior in performance to all former published works. 

\section{Acknowledgements}

We give thanks to Yichen Gong, Dawei Wang and Rong Zhang for sharing their pearls of wisdom. This work was supported by Shining Lab and Alibaba Group through Alibaba Research Intern Program.

\bibliographystyle{IEEEtran}

\bibliography{mybib}

\end{document}